\documentclass[sigconf]{acmart}
\usepackage{pdfpages}
\pdfoutput=1
\usepackage{booktabs} 

\copyrightyear{2019}
\acmYear{2019}
\setcopyright{iw3c2w3}
\acmConference[WWW '19 Companion]{Companion Proceedings of the 2019 World Wide Web Conference}{May 13--17, 2019}{San Francisco, CA, USA}
\acmBooktitle{Companion Proceedings of the 2019 World Wide Web Conference (WWW '19 Companion), May 13--17, 2019, San Francisco, CA, USA}
\acmPrice{}
\acmDOI{10.1145/3308560.3314195}
\acmISBN{978-1-4503-6675-5/19/05}

\usepackage{balance}

\usepackage{multirow}
\usepackage{booktabs}
\usepackage{mathtools}
\usepackage{subcaption}
\usepackage{amsfonts}
\usepackage{tikz}
\usepackage{pdflscape}
\usepackage{verbatim}
\usepackage{algorithm}
\usepackage{algorithmic}
\usepackage{afterpage}

\usetikzlibrary{arrows, positioning, shadows, calc,trees,backgrounds,shapes,decorations.pathmorphing,fit,positioning,chains}
\def\lav{gray!20}
\def\oran{gray!80}
\tikzstyle{peers}=[draw,circle,fill=\lav, \lav, minimum width=15pt, thin, align=center, anchor=base, text=black]
\tikzstyle{superpeers}=[draw, minimum height=1.6em, minimum width=1em, fill=\oran, double copy shadow={shadow xshift=1pt, shadow yshift=1pt, fill=white, draw}, draw, rectangle, \oran, thin, align=center, anchor=base, text=black]
\tikzstyle{legendsp}=[rectangle, draw, rounded corners, thin,fill=\oran, \oran, minimum width=1cm,align=center, anchor=base, text=black]
\tikzstyle{legendp}=[rectangle, draw, rounded corners, thin, fill=\lav, \lav, minimum width= 1cm,align=center, anchor=base, text=black]

\setcopyright{rightsretained}

\newcommand{\modelname}{\textit{GVNR}}
\newcommand{\modelnametext}{\textit{GVNR-t}}

\begin{document}

\title{Representation Learning for Recommender Systems with Application to the Scientific Literature}

\author{Robin Brochier}
\affiliation{%
  \institution{Universit\'e de Lyon, Lyon 2, ERIC EA3083 \\ Digital Scientific Research Technology}
  \city{Lyon}
  \state{France}
}
\email{robin.brochier@univ-lyon2.fr}

\renewcommand{\shortauthors}{R. Brochier}

\begin{abstract}
The scientific literature is a large information network linking various actors (laboratories, companies, institutions, etc.). The vast amount of data generated by this network constitutes a dynamic heterogeneous attributed network (HAN), in which new information is constantly produced and from which it is increasingly difficult to extract content of interest. In this article, I present my first thesis works in partnership with an industrial company, \textit{Digital Scientific Research Technology}. This later offers a scientific watch tool, \textit{Peerus}, addressing various issues, such as the real time recommendation of newly published papers or the search for active experts to start new collaborations. To tackle this diversity of applications, a common approach consists in learning representations of the nodes and attributes of this HAN and use them as features for a variety of recommendation tasks. However, most works on attributed network embedding pay too little attention to textual attributes and do not fully take advantage of recent natural language processing techniques. Moreover, proposed methods that jointly learn node and document representations do not provide a way to effectively infer representations for new documents for which network information is missing, which happens to be crucial in real time recommender systems. Finally, the interplay between textual and graph data in text-attributed heterogeneous networks remains an open research direction.                  
\end{abstract}

%
%



\maketitle

\section{Problem}

Many applications, which have become everyday tools, offer to search and filter the vast data sources available on the Web. In particular, there is a multitude of platforms dealing with scientific literature. From the simple search engine for scientific articles to the social network for researchers, all use, as data, the daily publications produced around the world. For the researcher facing this deluge of information, it has become difficult, if not impossible, to conduct a regular and exhaustive monitoring of his areas of expertise. The ongoing research presented in this paper, done in partnership with an industrial player \footnote{\textit{Digital Scientific Research Technology} (\textit{DSRT}) and its web application \textit{Peerus}: \url{https://peer.us}.}, deals with the problem of learning representations in heterogeneous networks of documents applied to the recommendation of scientific literature in real time.  

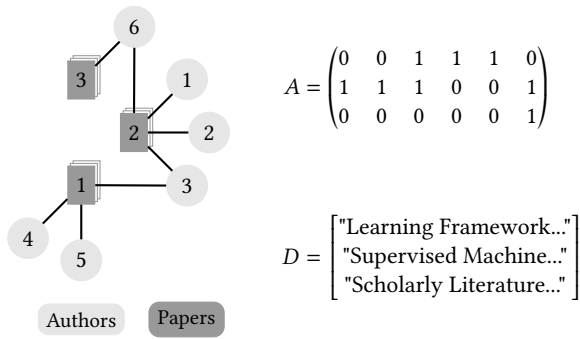
\begin{figure}[]
\centering
    \begin{subfigure}[b]{0.23\textwidth}
        \centering
        \begin{tikzpicture}[auto, thick, scale=0.35]
          \edef\mya{0}
          \foreach \place/\name in {{(0,-2)/a}, {(2,0)/b}, {(0,2)/d}}
           \pgfmathparse{int(\mya+1}
            \xdef\mya{\pgfmathresult}
            \node[superpeers] (\name) at \place {\mya};
           \foreach \pos/\i in {above right of/1, right of/2, below right of/3}
            \node[peers, \pos =b ] (b\i) {\i};
           \foreach \speer/\peer in {b/b1,b/b2,b/b3}
           \path (\speer) edge[-] (\peer);
           \path (a) edge[-] (b3);
           \node[peers, above right of=d] (d1){6};
           \path (d) edge[-] (d1);
           \path (b) edge[-] (d1);
           \edef\mya{3}
           \foreach \pos/\i in {below left of/1, below of/2}
            \pgfmathparse{int(\i+3)}
            \edef\mya{\pgfmathresult}
            \node[peers, \pos =a ] (a\i) {\mya};
           \foreach \speer/\peer in {a/a1,a/a2}
           \path (\speer) edge[-] (\peer);
           \node[legendsp] at (4,-7) {\small{Papers}};
           \node[legendp] at (0,-7.1) {\small{Authors}};
        \end{tikzpicture}
\end{subfigure}%
~~~~
\begin{subfigure}[b]{0.23\textwidth}
        \centering
        \begin{align*}
            A &= \begin{pmatrix}
              0        & 0       & 1       & 1         & 1        & 0\\
              1        & 1       & 1       & 0         & 0        & 1\\
              0        & 0       & 0       & 0         & 0        & 1\\
            \end{pmatrix}\\
            &\\
            &\\
            D &= \begin{bmatrix}
              \text{"Learning Framework..."} \\
              \text{"Supervised Machine..."}  \\
              \text{"Scholarly Literature..."}  \\
            \end{bmatrix} \\
            &
        \end{align*}
\end{subfigure}
\caption{Hypothetical example of scientific literature data, here constituted of 6 authors and 3 papers. The HAN can be noted $G=(V,E,D)$ with $V$ the nodes, $E$ the edges and $D$ the textual content of the papers. We note A the biadjacency matrix of the graph $G$.}
\label{fig:data}
\end{figure}

If the scientific information of a publication is mainly contained in the text that composes it, rich supplementary information is nested in its metadata. Thus, networks of co-authors, citations and places of publication of an article contain important information for the realization of a scientific recommender system. As such, the scientific literature constitutes a heterogeneous attributed network (HAN) and since new papers are constantly published, this HAN grows in real time (Figure \ref{fig:data} shows an hypothetical example). However, limited information might be observed from the new aggregated nodes. For example, a newly published paper hasn't incoming citation links and a PhD student has a limited number of past co-authorship links. To face this lack of information, the approach considered in the proposed research is to focus on learning strong representations of the attributes, particularly the textual contents of the articles, that can reflect the partially observable underlying network structure.  

\section{State of the art}

The quality and informativeness of data representation greatly influence the performance of machine learning algorithms. For this reason, a lot of efforts are devoted to devise new ways of learning representations \cite{bengio2013representation}. In Section \ref{soa:1}, I describe how the task of learning representations of nodes, \textit{i.e.} network embedding, is tightly connected to word embedding. Then, in Section \ref{soa:2}, I focus on the interplay between natural language processing and network embedding. Finally, in Section \ref{soa:3}, I present recent works that extend network embedding techniques to HAN.    

\subsection{From Word Embedding to Network Embedding} \label{soa:1}

The distributional hypothesis \cite{sahlgren2008distributional} forms the basis of word embedding algorithms. This assumes that the distributional similarity of words correlates strongly with their semantic similarity. In other words, if we learn a representation of a word that allows us to predict the other words that occurred in its context, we obtain a representation of its meaning.

Skip-Gram \cite{mikolov2013distributed} is an algorithm that builds representations of words by maximizing the log-likelihood of a multi-set of co-occurring word pairs. Skip-gram with Negative Sampling is a variation proposed in \cite{mikolov2013distributed} to effectively approach that log-likelihood. This is achieved by reducing the task to a classification which consists in distinguishing pairs of words that co-occur with false pairs that do not co-occur. An alternative approach, GloVe \cite{pennington2014glove}, learns representations of words by factoring a matrix of counts of co-occurrences of the words of a corpus. Its objective is to minimize the error of reconstruction of the matrix, considering only the non-zero values of co-occurrence counts.

Even though the distributional hypothesis originated in linguistics and is naturally leveraged for word embedding, Perozzi \textit{et al.} establish the connection with network embedding. To do so, they show that the frequency at which nodes appear in short random walks follows a power-law distribution, like the frequency of words in language \cite{perozzi2014deepwalk}. 
They propose DeepWalk, that consists in applying skip-gram with hierarchical softmax on a corpus of node sequences, deemed equivalent to sentences, generated with truncated random walks \cite{perozzi2014deepwalk}. For some specific tasks, the representations learned with DeepWalk offer large performance improvements. Thus, many subsequent works focus on modifying or extending DeepWalk. 
Node2vec replaces random walks with biased random walks, in order to better balance the exploration-exploitation trade-off, arguing that the added flexibility in exploring neighborhoods helps learning richer representations \cite{grover2016node2vec}.
VERSE \cite{tsitsulin2018verse} provides a scalable graph embedding algorithm by defining a versatile similarity matrix of the nodes and a learning algorithm using noise-contrastive estimation \cite{gutmann2010noise}, which provably converges to its objective in contrast to negative sampling. 

\subsection{Natural Language Processing in Networks of Documents} \label{soa:2}

As a special case of attributed networks, graphs of documents bring together the fields of natural language processing (NLP) and network embedding. A wide variety of unsupervised learning algorithms to represent words and documents have been proposed, from the well-known bag-of-word model \cite{harris1954distributional} to the recently introduced attention-based Transformer \cite{vaswani2017attention} adapted for unsupervised pre-training \cite{devlin2018bert}. However, few works have fully explored the interplay between NLP techniques and network embedding.    

NetPLSA \cite{mei2008topic} adapts a topic modelling algorithm by regularizing a statistical topic model with a harmonic regularizer based on a graph structure. It generates topics that reflect the underlying communities of the network, providing cleaner topics than regular statistical models.  

In \cite{yang2015network}, Yang \textit{et al.} prove that skip-gram with hierarchical softmax can be equivalently formulated as a matrix factorization problem. They then propose Text-Associated DeepWalk (TADW), to deal with networks of documents. TADW consists in constraining the factorization problem, with a pre-computed representation of documents via LSA \cite{deerwester1990ilsa}. As such, each node can be represented as the concatenation of a network embedding, and a projected textual embedding. 
CANE \cite{tu2017cane} aims to improve node representations in a structured corpus by applying a mutual attention mechanism over the textual contents associated with the vertices of a graph. Given a connected pair of nodes, the model produces textual representations for each node contextually to the other node. In this manner, there are as many representations for a single node as it has neighbors. This model has the advantage to produce interpretable weights for the words of a pair of documents, highlighting those that explain the network structure. 

\subsection{Heterogeneous Attributed Networks} \label{soa:3}

Real-world networks are often composed of several types of links and nodes which are associated with attributes. For example, the scientific literature is made of articles and authors, with directed citations links between papers, co-authorship links between authors and articles are associated with their textual content and their journal of publication. Many works have extended network embedding to handle heterogeneity and attributes in graphs. 

With Metapath2vec \cite{dong2017metapath2vec}, the authors propose to operate meta-path based random walks to handle heterogeneous nodes and links. These meta-path are hand-crafted schemes that guide a random walker over the network to generate nodes co-occurrences. Using a similar Skip-Gram based objective as DeepWalk, Metapath2vec achieves significant improvements on multi-class node classification and node clustering over traditional network embedding algorithms. 
\cite{huang2017label} introduces a Label informed Attributed Network Embedding (LANE) framework which jointly projects an attributed network and its labels into a unified embedding space by extracting their correlations. The mapping of the structural proximities in the attributed network and labels into an identical embedding space via correlation projections produces a significant improvement of the embeddings. 
Some works go beyond the factorization-based embedding approaches, introducing models that learn a function to generates embeddings by sampling and aggregating features from a node's local neighborhood. GraphSAGE \cite{hamilton2017inductive} makes use of learnable aggregator functions that allow to infer representations for unseen nodes, given their attributes and links. \cite{velickovic2018graph} adapt recent work on attention mechanism to compute a node representation by attending to its neighbors.

\section{Proposed Approach}

The proposed approach is intended to produce a novel model for learning representations of nodes and documents in a dynamic heterogeneous network with the goal to compute meaningful recommendations in real-time. The novelty results in the capacity of the model to infer representations of unseen documents for which no network information is available, in the same embedding space as the previously observed nodes. The approach is divided in three steps:
\begin{enumerate}
    \item design of a first model to learn node representations that can handle textual attributes. By opposition to TADW, which rely on previously learned LSA representations, this model should learn word and document embeddings from scratch to ease the inference for unseen documents. This step should validate the possibility to learn meaningful text representations from graph information only.
    \item improvement of the model by focusing on natural language processing. The model would be able to predict the similarity of documents in a network, based on their textual content only. It would deal with more advanced NLP techniques and further take advantage of the interplay between word and document representations and the network topology. Compared to CANE, the representations should be produced only from text information (still using the network as training supervision) and a strong emphasis on link prediction for unseen documents should be put.              
    \item integration of the heterogeneity to handle different types of nodes and links. This would allow to apply the model to a wider variety of tasks, such as user-item recommendation and expert finding. Handling the diversity of node and link types shouldn't rely on hand-crafted meta-path, as proposed in metapath2vec, but should be learned during the process similarly to GraphSAGE. At this point, the data provided by DSRT from \textit{Peerus} would serve as a strong online evaluation of the proposed model.  
\end{enumerate}

Step (1) has been achieved and is detailed in Section \ref{metho2}. The results are presented in Section \ref{results1}. More work on its theoretical background will be done in a near future. Step (2) is ongoing research, that I briefly present in Section \ref{metho3} and for which I provide some preliminary results in Section \ref{results2} motivating the research direction. For all evaluations, I detail the datasets used and the experimental setups in Section \ref{metho1}.   

\section{Methodology}

In this Section, I first provide an overview of the evaluations used for my research, then I detail a contribution corresponding to the first step of my thesis and I finally briefly address the planned methodology for the next steps.        

\subsection{Evaluation} \label{metho1}

I first detail some datasets commonly used in the literature. Then I briefly present traditional experiments conducted for evaluating network embedding. 

\subsubsection{Datasets} 

I present below two small datasets, Cora and CiteSeer, according to the treatments applied in \cite{sen2008collective} as well as a larger dataset, DBLP, widely used by the scientific community:
\begin{itemize}
\item \textbf{Cora} \cite{mccallum2000automating} is a network of scientific articles in the field of artificial learning, grouping 7 classes (scientific subdomains) with 2708 documents, 1433 distinct words in the vocabulary and 5429 citations links. 
\item \textbf{CiteSeer} \cite{giles1998citeseer} is a network of scientific articles grouping 6 classes on 3312 documents, 3703 distinct words in the vocabulary and 4732 citations links. 
\item \textbf{DBLP} \cite{ley2002dblp} is a database of several million scientific articles in the field of computer science, started in 1993. A powerful disambiguation system makes it possible to identify the authors \cite{ley2009dblp}.
\end{itemize}

Many other non-scientific datasets that present similar data structures exist. Among others, Q\&A websites and online encyclopedia provide rich sources of data for which we can tackle similar challenges as with the scientific literature. Moreover, the industrial player supporting these research provides a large dataset of scientific literature with user log activities that allows to apply and evaluate the proposed models to online recommendation tasks.   

\subsubsection{Experiments} 

To evaluate network representation learning models, it is common to use the nodes embeddings as input space for a linear algorithm to classify the nodes. For each set of representations produced by a particular algorithm, the proportion of learning representations is varied from 10\% to 50\% and the average prediction accuracy of the classifier is computed over the rest of the node representations, given a set of ground truth labels. This evaluation was used for the results in Section \ref{results1}. 

An extension of this evaluation consists in observing only from 30\% to 70\% of the nodes when learning the representations. Then, the accuracies of classification are computed on the unobserved nodes, using only their attributes for prediction. As such, we evaluate the algorithm on its capacity to infer representation for new unseen documents. This evaluation was used in Section \ref{results2}. Note that this is different from the inductive prediction performed by GraphSAGE, which also infers representations for unseen nodes, but with the knowledge of the actual new links of these new nodes.       

Finally, to evaluate the model of step (2), link prediction constitutes a good evaluation task. Several ways to generate a pair of training/test set exist (random, temporal). The goal is then to distinguish unseen links from non-existing ones. The most suited metric for this is the ROC AUC. The same way as with the previous classification task, it is possible to extend this evaluation for unobserved new documents by hiding a proportion of the nodes (and not of the links) during learning.    

\subsection{Document Network Embedding} \label{metho2}

In this section, I present the first contribution of my thesis \cite{brochier2019global}, \modelname{} (Global Vectors for Node Representation), a model to learn node representations with an extension, \modelnametext{}, to handle text-associated nodes. We seek to learn two sets of representations of the nodes $U \in \mathbb{R}^{n \times d}$ and $V \in \mathbb{R}^{n \times d}$, $n$ being the number of nodes in the network and $d$ the dimension of the learned embeddings.  

\subsubsection{Factorization Problem}

We formulate a factorization problem on a random-walk based co-occurrence counts matrix  $X$ generated from an input network, measuring the error of reconstruction only for positive coefficients and a fraction of randomly sampled zero coefficients:
\begin{equation}
\underset{U,V,b^U,b^V}{\mathrm{argmin}} \sum_{i=1}^n \sum_{j=1}^n s(x_{ij})\big(u_i \cdot v_j + b^U_i + b^V_j - \log (x_{ij})\big)^2
\end{equation}
$b^U_i$ and $b^V_j$ are two learned biases for the pair of node embeddings. The function $s$ effectively selects the coefficients considered for measuring the reconstruction error:
\begin{equation}
s(x_{ij}) = \begin{cases}
    		1 & \text{if } x_{ij} > 0,\\
    		m_{i} & \text{else, with } m_{i} \sim \text{Bernoulli}\Big(\frac{k\times n_i}{n-n_i}\Big).
    	\end{cases}
\end{equation}
It takes the value 1 for all positive coefficients of $X$, while for zero coefficients, its value is given by a Bernoulli random variable, $m_i$. It depends on the number of distinct nodes with which node $i$ co-occurs, $n_i$. We introduce a global hyper-parameter $k \in \mathbb{N^+_*}$, to control the proportion of zero coefficients incorporated into the reconstruction error.

\subsubsection{Extended Model for Networks of Documents}
\label{sec:extended_model}

In this brief section, we show how to extend \modelname{} to deal with networks where nodes are short text documents. 

Assuming word order is negligible for short documents (such as a scientific abstract), we can model them as bag-of-words and thus represent a document $j$ with a vector $\text{doc}_j \in \mathbb{N^+}^m$, $m$ being the size of the vocabulary. We can further assume that the meaning of a short text can be captured by averaging the representations of its words  \cite{le2014paragraph}.

Therefore, with $W \in \mathbb{R}^{m \times d}$ a word embedding matrix, we define the context-vector representation of a node in the following way: $v_j = \frac{\text{doc}_j ~ W}{|\text{doc}_j|_1}$ 

\subsection{Improving Document Network Embedding} \label{metho3}

\modelnametext{} is able to jointly learn word, document and node embeddings in a network of documents. However, the textual information could highly benefit from more recent works in the field of NLP. In this direction, the recently introduced Transformer as shown great promise in learning dependencies between words for text representation. Besides its low computational complexity and its strong results achieved on neural machine translation and unsupervised pre-training for language understanding, its core unit, the Scaled Dot-Product Attention, provides a good basis for extending \modelnametext{}{}. 

The Scaled Dot-Product Attention takes as input, a set of keys $K$ and values $V$, corresponding to projected representations of words in a documents, and a query $q$, possibly any kind of vector lying in the same space as the keys. As output, it generates a weighted sum of the values, whose weights are produced by confronting the keys with the query, following the formula: $\text{Attention}(q,K,V) = \text{softmax}(\frac{qK^T}{\sqrt{d_k}})V$, $d_k$ being the dimension of the query and the keys.  

My current research focus on exploring the use of this attention mechanism for mutual attention between pairs of documents in a network. Using pre-trained word embeddings, I try to find a suitable variation of this unit for generating sparse weights (hence using other functions than the softmax) and I explore several ways to build an efficient query $q$ for mutual attention.    

\section{Results} \label{results}

I first present the results obtained on multi-class classification by \modelname{} and its extension with text and then I show some preliminary results indicating that more emphasis should be set on the representations of the textual content of nodes in a network. Finally, I show an example of a visualization of the weights learned by a preliminary model adapting the Scaled Dot-Product Attention for networks of documents.    

\subsection{Results for \modelname{} and it Extension With Text} \label{results1}

The results presented in Table \ref{citation1text} show the average accuracies for multi-class classification obtained on Cora. First, we observe that \modelname{} produces competitive representations with DeepWalk. Its extension \modelnametext{}, with the integration of the textual content of the documents, significantly improves the quality of the embeddings, achieving even better performances than TADW which relies however on textual representations more time-consuming, obtained with latent semantic analysis.

\begin{table}[h]
\center
\caption{Results of multi-class classification on Cora.}
\begin{tabular}{l|ccccc}
\% of training data &10\%    &20\%    &30\%    &40\%    &50\%   \\ \hline
DeepWalk &67.8   &71.6   &74.5   &75.8   &79.2 \\
\modelname{} ($x_{\text{min}}=1$) &\textbf{69.5}  &\textbf{72.6}   &\textbf{75.9}   &\textbf{78.1}   &\textbf{80.2} \\ \hline
DeepWalk+LSA &73.8   &77.9   &78.4   &78.1   &78.1 \\ 
TADW &77.1   &78.8   &78.2   &78.8   &78.6 \\ 
\modelnametext{} &\textbf{79.3}   &\textbf{80.7}   &\textbf{80.8}   &\textbf{81.4}   &\textbf{81.1} \\ \hline
TADW (text only) &60.5    &69.3    &72.7    &73.6    &74.5     \\ 
\modelnametext{} (text only) &\textbf{74.5}    &\textbf{76.5}    &\textbf{78.5}   &\textbf{78.6}    &\textbf{79.8}   \\
\end{tabular}
\label{citation1text}
\end{table}

\subsection{Motivation for a Stronger NLP Component} \label{results2} 

To gain more insight into the textual representations that are learned, Table \ref{citation1text} shows the accuracies obtained by TADW and \modelnametext{} with their respective text components only. We see that \modelnametext{} has significantly higher accuracies than TADW, but it is unclear if this is due to an underlying better natural language understanding. Table \ref{citation1unseen} shows the results of classification when predicting on unseen documents. We observe that if \modelnametext{} is capable of generalizing on the text attributes of the nodes, TADW relatively fails. However, the results achieved by \modelnametext{} are still lower than expected and motivates the use of more advanced NLP techniques to achieve better generalization.  

\begin{table}[h]
\center
\caption{Unseen documents classification accuracies on Cora.}
\begin{tabular}{l|ccccc}
\% of training data        &30\%    &40\%    &50\%      &60\%    &70\%     \\ \hline
TADW &39.7      &48.9   &50.2   &51.5   &52.2  \\
\modelnametext{}    &\textbf{64.3}   &\textbf{67.7}   &\textbf{71.2}   &\textbf{73.6}   &\textbf{73.8}  \\\hline
\end{tabular}
\label{citation1unseen}
\end{table}

\subsection{Mutual Attention for Networks of Documents} \label{results2} 
\begin{figure}[h]
   \includegraphics[scale=1.65]{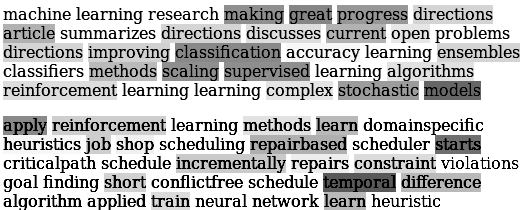}
   \caption{\label{attention} Mutual attention weights for a pair of documents extracted from Cora.}
\end{figure}

My ongoing research is meant to adapt the Scaled Dot-product attention mechanism for network of documents. The hope is to find a way to effectively infer weights for the words of the documents that strongly support (i.e highlight evidence for) the links in the network. Figure \ref{attention} shows an example of generated weights with a first draft of such a model. Interestingly, the model highlights words related to the field of reinforcement learning in both texts.

\section{Conclusion and Future Work}

Text data and network data are the two most represented information types on the World Wide Web. Building meaningful representations for both is a crucial step for the design of efficient recommender systems. Particularly, the ever growing scientific literature constitute a dynamic heterogeneous text-attributed network. The interplay between the textual content of scientific publications and the networks dynamics of the actors of the research brings strong challenges. 

The proposed research aims at discovering an efficient model to represent the variety of nodes and links in the scientific HAN in a unique representation space in order to tackle a wide variety of recommendation tasks. The first works achieved during that research allowed to validate the complementarity of the two sources of information, text and graph, to learn meaningful representations. Ongoing research now aims at improving the natural language understanding component of the model, to truly be able to generate representations for streams of new documents. A last step will focus on extending the coverage of the types of nodes and links that the model can handle and intensively evaluating it on a wide variety of real-world recommendation tasks.         

\bibliographystyle{ACM-Reference-Format}
\balance 
\bibliography{main}

\end{document}